\documentclass[11pt,a4paper]{article}
\usepackage[hyperref]{naaclhlt2019}
\usepackage{times}
\usepackage{latexsym}
\usepackage{textcomp}
\usepackage{siunitx}
\usepackage{amsmath}
\usepackage{booktabs}
\usepackage{multirow}
\usepackage{graphicx}
\usepackage{url}
\usepackage{footnote}
\usepackage[symbol,para]{footmisc}
\usepackage{xcolor}
\usepackage{xspace}
\usepackage{float}
\usepackage{placeins}
\usepackage{subcaption}
\usepackage{balance}

\aclfinalcopy % Uncomment this line for the final submission

\newcommand{\parheader}[1]{{\smallskip \noindent \bf #1.}}

\newcommand{\FOne}{$\text{F}_1$}
\definecolor{hilite}{RGB}{0, 100, 200}

\newcommand{\BLSTM}[1]{LSTM$_{base}$}
\newcommand{\BLSTMR}[1]{LSTM$_{reg}$}
\newcommand{\BERTL}[1]{BERT$_{large}$}
\newcommand{\BERTB}[1]{BERT$_{base}$}
\defcitealias{yang2018sgm}{Yang et al.'s (2018)}

\title{DocBERT: BERT for Document Classification}

\author{Ashutosh Adhikari, Achyudh Ram, Raphael Tang, \and Jimmy Lin\vspace{0.1cm}\\
David R. Cheriton School of Computer Science\\
University of Waterloo\\
{\tt \{adadhika, arkeshav, r33tang, jimmylin\}@uwaterloo.ca}}

\date{}

\begin{document}
\maketitle
\begin{abstract}
We present, to our knowledge, the first application of BERT to document classification.
A few characteristics of the task might lead one to think that BERT is not the most appropriate model:\ syntactic structures matter less for content categories, documents can often be longer than typical BERT input, and documents often have multiple labels.
Nevertheless, we show that a straightforward classification model using BERT is able to achieve the state of the art across four popular datasets.
To address the computational expense associated with BERT inference, we distill knowledge from \BERTL{} to small bidirectional LSTMs, reaching \BERTB{} parity on multiple datasets using 30$\times$ fewer parameters.
The primary contribution of our paper is improved baselines that can provide the foundation for future work.
\end{abstract}

\section{Introduction}
Until recently, the dominant paradigm in approaching natural language processing (NLP) tasks has been to concentrate on neural architecture design, using only task-specific data and word embeddings such as GloVe~\cite{pennington2014glove}.
The NLP community is, however, witnessing a dramatic paradigm shift toward the pre-trained deep language representation model, which achieves the state of the art in question answering, sentiment classification, and similarity modeling, to name a few.
Bidirectional Encoder Representations from Transformers~(BERT; \citealp{devlin2018bert}) represents one of the latest developments in this line of work.
It outperforms its predecessors, ELMo~\cite{elmo} and GPT~\cite{radfordimproving}, by a wide margin on multiple NLP tasks.

This approach consists of two stages: first, BERT is pre-trained on vast amounts of text, with an unsupervised objective of masked language modeling and next-sentence prediction.
Next, this pre-trained network is then fine-tuned on task-specific, labeled data.

BERT, however, has not yet been fine-tuned for document classification.
Why is this worth exploring?
For one, modeling syntactic structure has been arguably less important for document classification than for typical BERT tasks such as natural language inference and paraphrasing.
This claim is supported by our observation that logistic regression and support vector machines are exceptionally strong document classification baselines.
For another, documents often have multiple labels across many classes, which is again uncharacteristic of the tasks that BERT examines.

In this paper, we first describe fine-tuning BERT for document classification to establish state-of-the-art results on four popular datasets.
This increase in model quality, however, comes at a heavy computational expense.
BERT contains hundreds of millions of parameters, while the previous baseline uses less than four million and performs inference forty times faster.

To alleviate this computational burden, we apply knowledge distillation~\cite{hintonkd} to transfer knowledge from \BERTL{}, the large BERT variant, to the previous, much smaller state-of-the-art BiLSTM.
As a result of this procedure, with a few additional tricks for effective knowledge transfer, we achieve results comparable to \BERTB{}, the smaller BERT variant, using a model with 30$\times$ fewer parameters.

Our contributions in this paper are two fold:\
First, we establish state-of-the-art results for document classification by simply fine-tuning BERT;
Second, we demonstrate that BERT can be distilled into a much simpler neural model that provides competitive accuracy at a far more modest computational cost.

\section{Background and Related Work}

Over the last few years, neural network-based architectures have dominated the task of document classification.
\citet{liu2017deep} develop XMLCNN for addressing this problem's multi-label nature, which they call extreme classification.
XMLCNN is based on the popular KimCNN~\cite{kim2014convolutional}, except with wider convolutional filters, adaptive dynamic max-pooling~\cite{chendynamicpool, Johnsondynamicpool}, and an additional bottleneck layer to better capture the features of large documents.
Another popular model, Hierarchical Attention Network (HAN; \citealp{yang2016hierarchical}) explicitly models hierarchical information from documents to extract meaningful features, incorporating word-~and sentence-level encoders (with attention) to classify documents.
\citet{yang2018sgm} propose a generative approach for multi-label document classification, using encoder--decoder sequence generation models (SGMs) for generating labels for each document.
Contrary to the previous papers, \citet{adhikari2019rethinking} propose \BLSTMR{}, a simple, properly-regularized single-layer BiLSTM, which represents the current state of the art.

In the current paradigm of pre-trained models, methods like BERT~\cite{devlin2018bert} and XLNet~\cite{Yang2019XLNetGA} have been shown to achieve the state of the art in a variety of tasks including question answering, named entity recognition, and natural language inference.
However, these models have a prohibitively large number of parameters and require substantial computational resources, even to carry out a single inference pass.
Similar concerns related to high inference latency or heavy run-time memory requirements have led to a myriad of works, such as error-based weight-pruning \cite{lecunprune}, and more recently, model sparsification and channel pruning \cite{loprune, netslim}.

Knowledge distillation (KD; \citealp{bakd, hintonkd}) has been shown to be an effective compression technique which ``distills'' information learned by a larger model (the teacher) into a smaller model (the student).
KD uses the class probabilities produced by a pre-trained teacher, the soft targets, to train a student model over a transfer set (the examples over which distillation takes place).
Being model agnostic, the approach is suitable for our study, as it enables the transfer of knowledge between different types of architectures, unlike most of the other model compression techniques.

\section{Our Approach}

To adapt \BERTB{} and \BERTL{} models for document classification, we follow \citet{devlin2018bert} and introduce a fully-connected layer over the final hidden state corresponding to the \texttt{[CLS]} input token.
During fine-tuning, we optimize the entire model end-to-end, with the additional softmax classifier parameters $W \in {\rm I\!R}^{K \times H}$, where $H$ is the dimension of the hidden state vectors and $K$ is the number of classes.
We minimize the cross-entropy and binary cross-entropy loss for single-label and multi-label tasks, respectively.

Next, we distill knowledge from the fine-tuned \BERTL{} into the much smaller \BLSTMR{}, which represents the previous state of the art~\cite{adhikari2019rethinking}.
We perform KD by using the training examples, along with minor augmentations to form the transfer set.
In accordance with \citet{hintonkd}, we combine the two objectives of classification using the target labels ($\mathcal{L}_{classification}$) and distillation ($\mathcal{L}_{distill}$) using the soft targets, for each example of the transfer set.

We use the target labels to minimize the standard cross-entropy or binary cross-entropy loss depending on the type of dataset (multi-label or single-label).
For the distillation objective, we minimize the Kullback--Leibler (KL) divergence $\textrm{KL}(p || q)$ where $p$ and $q$ are the class probabilities produced by the student and the teacher models, respectively. We define the final objective as:
\begin{equation}
    \label{eq:loss}
    \mathcal{L} = \mathcal{L}_{classification} + \lambda \cdot \mathcal{L}_{distill}
\end{equation}
where $\lambda \in R$  weighs the losses' contributions to the final objective.

\section{Experimental Setup}

\renewcommand*{\thefootnote}{\arabic{footnote}}

Using Hedwig,\footnote{\url{https://github.com/castorini/hedwig}} an open-source deep learning toolkit with a number of implementations of document classification models, we compare the fine-tuned BERT models against HAN, KimCNN, XMLCNN, SGM, and \BLSTMR{}.
For simple yet competitive baselines, we run the default logistic regression (LR) and support vector machine (SVM) implementations from Scikit-Learn~\cite{scikit-learn}, trained on the tf--idf vectors of the documents.

We use Nvidia Tesla V100 and P100 GPUs for fine-tuning BERT and run the rest of the experiments on RTX 2080 Ti and GTX 1080 GPUs.
We use PyTorch 0.4.1 as the backend framework, and Scikit-learn 0.19.2 for computing the tf--idf vectors and implementing LR and SVMs.

\begin{table}
\centering
\begin{tabular}{lrrrr}
\toprule[1pt]
Dataset & $C$ & $N$ & $W$ & $S$ \\
\midrule
Reuters & 90 & 10,789 & 144.3 & 6.6 \\
AAPD & 54 & 55,840 & 167.3 & 1.0 \\
IMDB & 10 & 135,669 & 393.8 & 14.4 \\
Yelp 2014 & 5 & 1,125,386 & 148.8 & 9.1 \\ \bottomrule[1pt]
\end{tabular}
\caption{Summary of the datasets. $C$ denotes the number of classes in the dataset, $N$ the number of samples, and $W$ and $S$ the average number of words and sentences per document, respectively.}
\label{table:datasets}
\end{table}

\subsection{Datasets} \label{sec:datasets}

We use the following four datasets to evaluate BERT:\ Reuters-21578 (Reuters; \citealp{apte1994automated}), arXiv Academic Paper dataset (AAPD; \citealp{yang2018sgm}), IMDB reviews, and Yelp 2014 reviews.
Reuters and AAPD are multi-label datasets while documents in IMDB and Yelp '14 contain only a single label.

For Reuters, we use the standard ModApt\'{e} splits~\cite{apte1994automated}; for AAPD, we use the splits provided by \citet{yang2018sgm}; for IMDB and Yelp, following \citet{yang2016hierarchical}, we randomly sample 80\% of the data for training and 10\%  each for validation and test.

We summarize the statistics of the datasets used in our study in Table \ref{table:datasets}.

\subsection{Training and Hyperparameters}

While fine-tuning BERT, we optimize the number of epochs, batch size, learning rate, and maximum sequence length (MSL), the number of tokens that documents are truncated to.
We observe that model quality is quite sensitive to the number of epochs, and thus the setting must be tailored for each dataset.
We train on Reuters, AAPD, and IMDB for 30, 20, and 4 epochs, respectively.
Due to resource constraints, we train on Yelp for only one epoch.
As is the case with \citet{devlin2018bert}, we find that choosing a batch size of 16, learning rate of 2$\times 10^{-5}$, and MSL of 512 tokens yields optimal performance on the validation sets of all datasets.
More details are provided in the appendix.

For distillation, we train the \BLSTMR{} model to capture the learned representations from \BERTL{} using the objective shown in Equation~(\ref{eq:loss}).
We use a batch size of 128 for the multi-label tasks and 64 for the single-label tasks.
We find the learning rates and dropout rates used in \citet{adhikari2019rethinking} to be optimal even for the distillation process.

To build an effective transfer set for distillation as suggested by \citet{hintonkd}, we augment the training splits of the datasets by applying POS-guided word swapping and random masking~\cite{ralphdistill}.
The transfer set sizes for Reuters, IMDB and AAPD are 3$\times$, 4$\times$, and 4$\times$ their training splits respectively, whereas only 1$\times$ (i.e., no data augmentation) the corresponding training split for Yelp2014 due to computational restrictions.
We use a $\lambda$ of 1 for the multi-label datasets and 4 for the single-label datasets.

\begin{table*}
\setlength{\tabcolsep}{4.25pt}
\centering
\scalebox{0.89}{
\begin{tabular}{@{}rlcccccccc@{}}
\toprule[1pt]
\multirow{2}{*}{\textbf{\#}} &
\multirow{2}{*}{\textbf{Model}} & \multicolumn{2}{c}{\textbf{Reuters}} & \multicolumn{2}{c}{\textbf{AAPD}} & \multicolumn{2}{c}{\textbf{IMDB}} & \multicolumn{2}{c}{\textbf{Yelp '14}} \\ \cmidrule(lr){3-4} \cmidrule(lr){5-6} \cmidrule(lr){7-8} \cmidrule(lr){9-10}
 & & \multicolumn{1}{l}{Val. \FOne} & \multicolumn{1}{l}{Test \FOne} & \multicolumn{1}{l}{Val. \FOne} & \multicolumn{1}{l}{Test \FOne} & \multicolumn{1}{l}{Val. Acc.} & \multicolumn{1}{l}{Test Acc.} & \multicolumn{1}{l}{Val. Acc.} & \multicolumn{1}{l}{Test Acc.} \\ \midrule
 1 & LR & \multicolumn{1}{l}{77.0} & \multicolumn{1}{l}{74.8} & \multicolumn{1}{l}{67.1} & \multicolumn{1}{l}{64.9} & \multicolumn{1}{l}{43.1} & \multicolumn{1}{l}{43.4} & \multicolumn{1}{l}{61.1} & \multicolumn{1}{l}{60.9} \\
 2 & SVM & \multicolumn{1}{l}{89.1} & \multicolumn{1}{l}{86.1} & \multicolumn{1}{l}{71.1} & \multicolumn{1}{l}{69.1} & \multicolumn{1}{l}{42.5} & \multicolumn{1}{l}{42.4} & \multicolumn{1}{l}{59.7} & \multicolumn{1}{l}{59.6} \\
3 & KimCNN Repl. & \multicolumn{1}{l}{83.5 \textpm 0.4} & \multicolumn{1}{l}{80.8 \textpm 0.3} & \multicolumn{1}{l}{54.5 \textpm 1.4} & \multicolumn{1}{l}{51.4 \textpm 1.3} & \multicolumn{1}{l}{42.9 \textpm 0.3} & \multicolumn{1}{l}{42.7 \textpm 0.4} & \multicolumn{1}{l}{66.5 \textpm 0.1} & \multicolumn{1}{l}{66.1 \textpm 0.6} \\
4 & KimCNN Orig. & \multicolumn{1}{l}{--} & \multicolumn{1}{l}{--} & \multicolumn{1}{l}{--} & \multicolumn{1}{l}{--} & \multicolumn{1}{l}{--} & \multicolumn{1}{l}{37.6\footnotemark[8]} & \multicolumn{1}{l}{--} & \multicolumn{1}{l}{61.0\footnotemark[8]} \\
5 & XML-CNN Repl. & \multicolumn{1}{l}{88.8 \textpm 0.5} & \multicolumn{1}{l}{86.2 \textpm 0.3} & \multicolumn{1}{l}{70.2 \textpm 0.7} & \multicolumn{1}{l}{68.7 \textpm 0.4} & \multicolumn{1}{l}{--} & \multicolumn{1}{l}{--} & \multicolumn{1}{l}{--} & \multicolumn{1}{l}{--} \\
6 & HAN Repl. & \multicolumn{1}{l}{87.6 \textpm 0.5} & \multicolumn{1}{l}{85.2 \textpm 0.6} & \multicolumn{1}{l}{70.2 \textpm 0.2} & \multicolumn{1}{l}{68.0 \textpm 0.6} & \multicolumn{1}{l}{51.8 \textpm 0.3} & \multicolumn{1}{l}{51.2 \textpm 0.3} & \multicolumn{1}{l}{68.2 \textpm 0.1} & \multicolumn{1}{l}{67.9 \textpm 0.1} \\
7 & HAN Orig. & \multicolumn{1}{l}{--} & \multicolumn{1}{l}{--} & \multicolumn{1}{l}{--} & \multicolumn{1}{l}{--}  & \multicolumn{1}{l}{--} & \multicolumn{1}{l}{49.4\footnotemark[3]} & \multicolumn{1}{l}{--} & \multicolumn{1}{l}{{70.5}\footnotemark[3]} \\
8 & SGM Orig. & \multicolumn{1}{l}{82.5 \textpm 0.4} & \multicolumn{1}{l}{78.8 \textpm 0.9} & \multicolumn{1}{l}{--} & \multicolumn{1}{l}{71.0\footnotemark[2]} & \multicolumn{1}{l}{--} & \multicolumn{1}{l}{--} & \multicolumn{1}{l}{--} & \multicolumn{1}{l}{--} \\
9 & \BLSTMR{} & \multicolumn{1}{l}{89.1 \textpm 0.8} & \multicolumn{1}{l}{87.0 \textpm 0.5} & \multicolumn{1}{l}{73.1 \textpm 0.4} & \multicolumn{1}{l}{70.5 \textpm 0.5} & \multicolumn{1}{l}{53.4 \textpm 0.2} & \multicolumn{1}{l}{52.8 \textpm 0.3} & \multicolumn{1}{l}{69.0 \textpm 0.1} & \multicolumn{1}{l}{68.7 \textpm 0.1} \\
\midrule
10 & \BERTB{} & \multicolumn{1}{l}{90.5} & \multicolumn{1}{l}{89.0}  & \multicolumn{1}{l}{75.3}  & \multicolumn{1}{l}{73.4} & \multicolumn{1}{l}{54.4} & \multicolumn{1}{l}{54.2} & \multicolumn{1}{l}{72.1} & \multicolumn{1}{l}{72.0} \\
11 & \BERTL{} & \multicolumn{1}{l}{\textbf{92.3}}  & \multicolumn{1}{l}{\textbf{90.7}}  & \multicolumn{1}{l}{\textbf{76.6}}  & \multicolumn{1}{l}{\textbf{75.2}}  & \multicolumn{1}{l}{\textbf{56.0}} & \multicolumn{1}{l}{\textbf{55.6}}  & \multicolumn{1}{l}{\textbf{72.6}}  & \multicolumn{1}{l}{\textbf{72.5}}  \\
\midrule
12 & KD-\BLSTMR{} & \multicolumn{1}{l}{91.0 \textpm 0.2}  & \multicolumn{1}{l}{88.9 \textpm 0.2} & \multicolumn{1}{l}{75.4 \textpm 0.2} & \multicolumn{1}{l}{72.9  \textpm 0.3}  & \multicolumn{1}{l}{54.5 \textpm 0.1}  & \multicolumn{1}{l}{53.7 \textpm 0.3} & \multicolumn{1}{l}{69.7  \textpm 0.1}  & \multicolumn{1}{l}{69.4 \textpm 0.1}  \\
\bottomrule[1pt]
\end{tabular}}
\caption{\renewcommand*{\thefootnote}{\fnsymbol{footnote}}Results for each model on the validation and test sets. Best values are bolded. \textit{Repl}.\ reports the mean of five runs from our reimplementations; \textit{Orig.}\ refers to point estimates from \footnotemark[2]\citet{yang2018sgm}, \footnotemark[3]\citet{yang2016hierarchical}, and \footnotemark[8]\citet{tang2015document}. KD-\BLSTMR{} represents the distilled \BLSTMR{} using the fine-tuned \BERTL{}.\renewcommand*{\thefootnote}{\arabic{footnote}}}
\label{table:results}
\end{table*}

\begin{table}[t]
\centering
\scalebox{1.0}{
\begin{tabular}{lrr}
\toprule[1pt]
Dataset & \BLSTMR{} & \BERTB{}  \\
\midrule
Reuters & 0.5 (1$\times$)  & 30.3 (60$\times$)    \\
AAPD & 0.3 (1$\times$) & 15.8 (50$\times$)  \\
IMDB & 6.8 (1$\times$) & 243.6 (40$\times$) \\
Yelp'14 & 20.6 (1$\times$)  & 1829.9 (90$\times$) \\ \bottomrule[1pt]
\end{tabular}}
\caption{Comparison of inference latencies (seconds) on validation sets with batch size 128.}
\label{table: inference}
\end{table}

\section{Results and Discussion}
We report the mean \FOne{} scores for multi-label datasets and accuracy for single-label datasets, along with the corresponding standard deviation, across five runs in Table \ref{table:results}.
We copy values for rows 1--9 from \citet{adhikari2019rethinking}.
Due to resource limitations, we report the scores from only a single run for \BERTB{} and \BERTL{}.

\begin{figure*}[t]
    \centering
       	\includegraphics[width=0.49\linewidth]{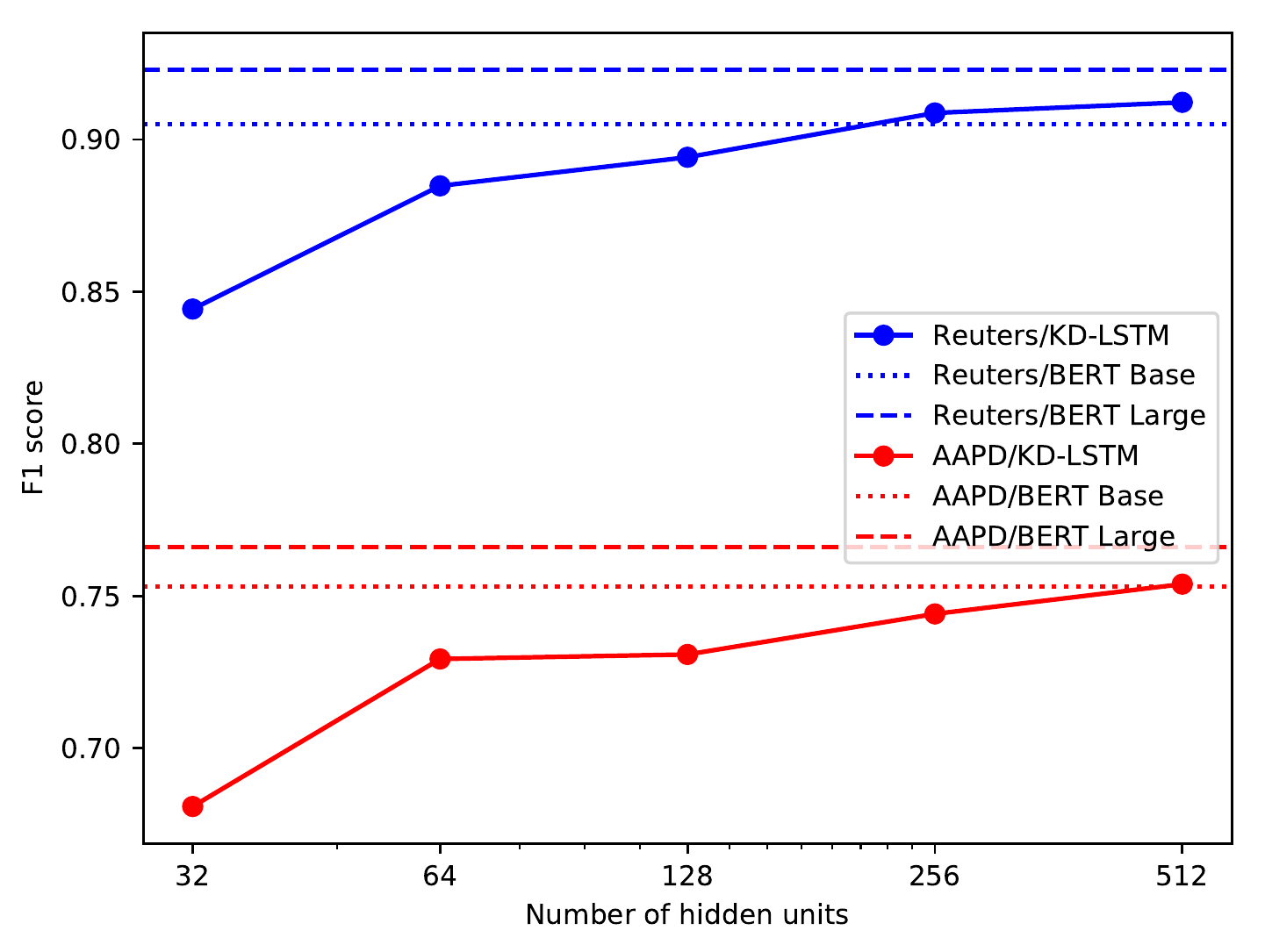}
    	\includegraphics[width=0.49\linewidth]{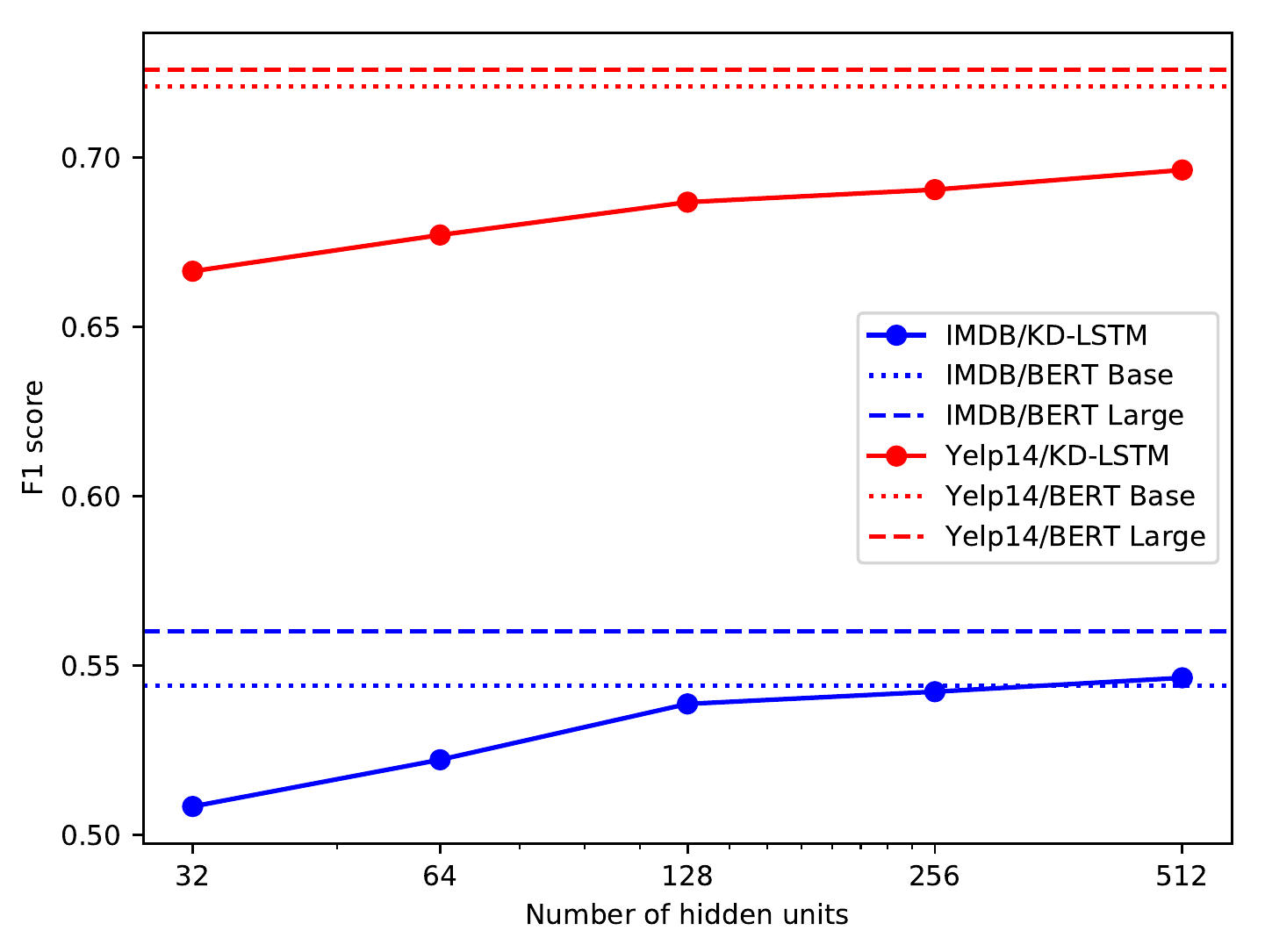}

    \caption{Effectiveness of KD-\BLSTMR{} vs.\ \BERTB{} and \BERTL{}}
    \label{fig:performance}
\end{figure*}

Consistent with \citet{devlin2018bert}, \BERTL{} achieves state-of-the-art results on all four datasets, followed by \BERTB{} (see Table \ref{table:results}, rows 10 and 11).
The considerably simpler \BLSTMR{} model (row 9) achieves high scores, coming close to the quality of \BERTB{}.
However, it is worth noting that all of the models above row 10 take only a fraction of the time and memory required for training the BERT models.

Surprisingly, distilled \BLSTMR{} (KD-\BLSTMR{}, row 12) achieves parity with \BERTB{} on average for Reuters, AAPD, and IMDB.
In fact, it outperforms \BERTB{} (on both dev and test) in at least one of the five runs.
For Yelp, we see that KD-\BLSTMR{} reduces the difference between \BERTB{} and \BLSTMR{}, but not to the same extent as in the other datasets.

To put things in perspective, Table \ref{table: inference} reports the inference times on the validation sets of all the datasets.
We calculate the inference times with batch size 128 for all the datasets on a single RTX 2080 Ti.
The relative speedup achieved by KD-\BLSTMR{} is at least around 40$\times$ with respect to \BERTB{}.
Additionally, Figure~\ref{fig:performance} shows the comparison between the number of parameters and prediction quality on the validation sets.
These plots convey the effectiveness of the KD-\BLSTMR{} model with different numbers of hidden units:\ 32, 64, 128, 256, and 512.
We find that KD-\BLSTMR{}, with just 256 hidden units (i.e., $\sim 1\%$ parameters of \BERTB{}) attains parity with \BERTB{} on Reuters, while for AAPD, 512 hidden units ($\sim 3\%$ parameters of \BERTB{}) are enough to overtake \BERTB{}.

\section{Conclusion and Future Work}

In this paper we improve the baselines for document classification by fine-tuning BERT.
We also use the knowledge learned by BERT models to improve the effectiveness of a single-layered light-weight BiLSTM model, \BLSTMR{}, using knowledge distillation.
In fact, we show that the distilled \BLSTMR{} model achieves \BERTB{} parity on a majority of datasets, resulting in over 30$\times$ compression in terms of the number of parameters and at least 40$\times$ faster inference times.

For future work, it would be interesting to study the effects of distillation over a range of neural-network architectures.
Alternatively, formulating specific model compression techniques in the context of transformer models deserves exploration.

\section*{Acknowledgments}

This research was supported by the Natural Sciences and Engineering Research Council (NSERC) of Canada, and enabled by computational resources provided by Compute Ontario and Compute Canada.

\bibliography{emnlp2019}
\bibliographystyle{acl_natbib}

\clearpage

\appendix

\section{Appendix} \label{appendix}

\subsection{Hyperparameter Analysis}

\begin{figure}[H]
\centering
\includegraphics[scale=0.3]{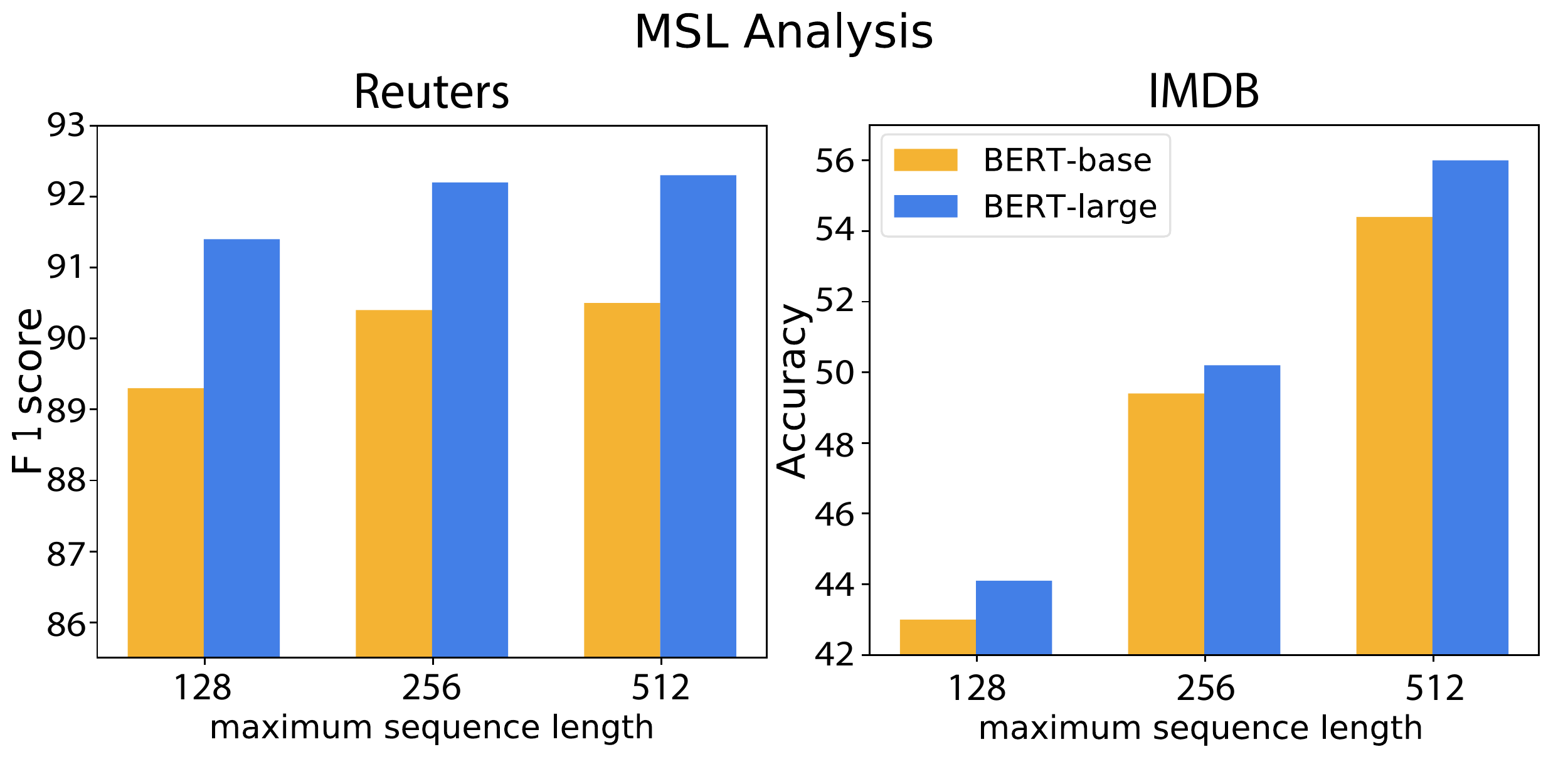}
\includegraphics[scale=0.3]{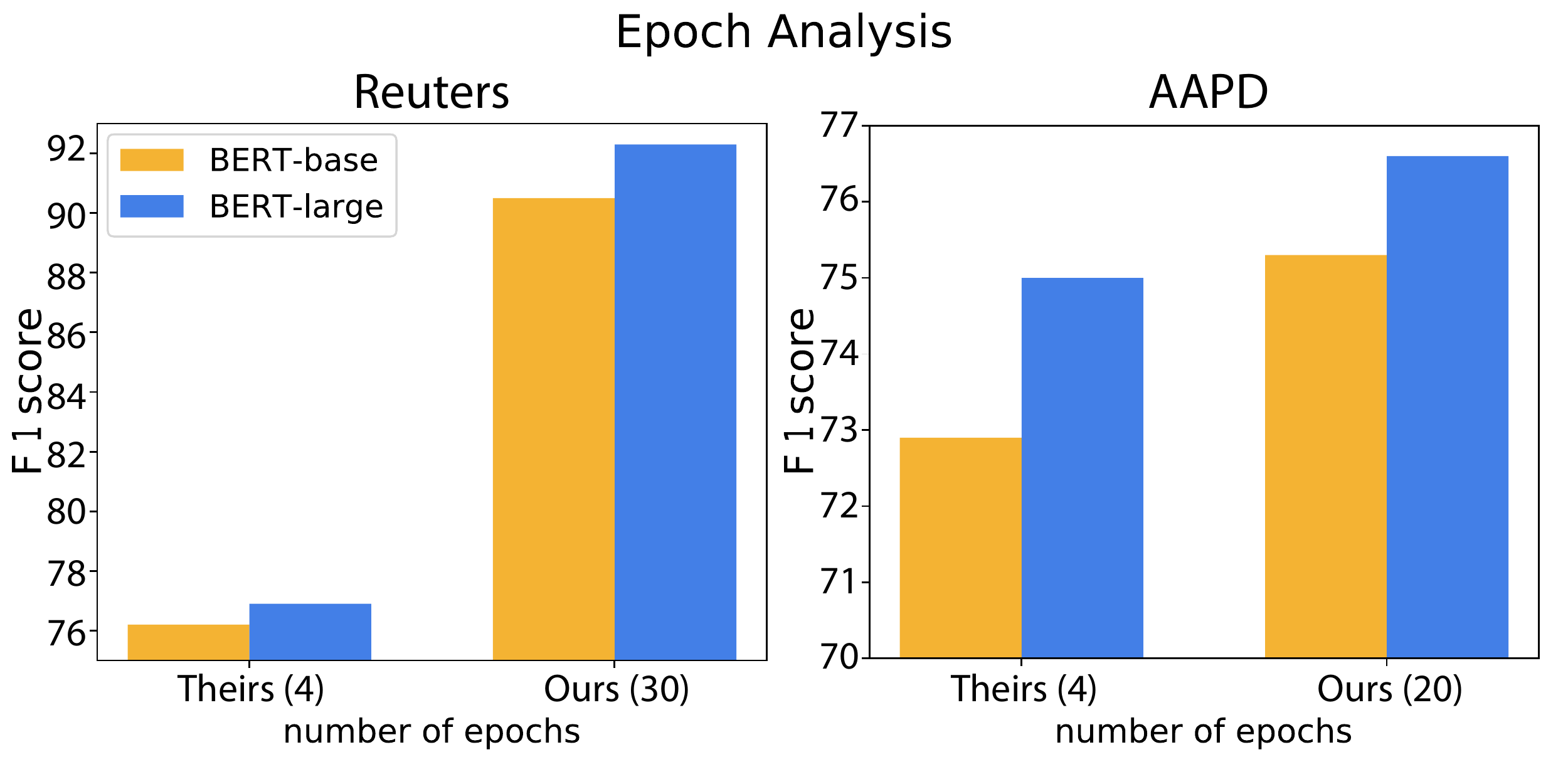}
\caption{Results on the validation set from varying the MSL and the number of epochs.}
\label{fig:ablation}
\end{figure}

\parheader{MSL analysis}
A decrease in the maximum sequence length (MSL) corresponds to only a minor loss in \FOne{} on Reuters (see top-left subplot in Figure~\ref{fig:ablation}), possibly due to Reuters having shorter documents.
On IMDB (top-right subplot in Figure~\ref{fig:ablation}), lowering the MSL corresponds to a drastic fall in accuracy, suggesting that the entire document is necessary for this dataset.

On the one hand, these results appear obvious.
Alternatively, one can argue that, since IMDB contains longer documents, truncating tokens may hurt less.
The top two subplots in Figure~\ref{fig:ablation} show that this is \textit{not} the case, since truncating to even 256 tokens causes accuracy to fall lower than that of the much smaller LSTM$_{reg}$ (see Table~\ref{table:results}).
From these results, we conclude that any amount of truncation is detrimental in document classification, but the level of degradation may differ.

\parheader{Epoch analysis}
The bottom two subplots in Figure~\ref{fig:ablation} illustrate the \FOne{} score of BERT fine-tuned using different numbers of epochs for AAPD and Reuters.
Contrary to \citet{devlin2018bert}, who achieve the state of the art on small datasets with only a few epochs of fine-tuning, we find that smaller datasets require many more epochs to converge.
On both the datasets (see Figure~\ref{fig:ablation}), we see a significant drop in model quality when the BERT models are fine-tuned for only four epochs, as suggested in the original paper.
On Reuters, using four epochs results in an \FOne{} worse than even logistic regression (Table~\ref{table:results}, row 1).

\end{document}